\documentclass[utf8]{FrontiersinHarvard}
\usepackage{url,hyperref,microtype,subcaption}
\usepackage{amsmath}
\usepackage{mathpazo}

\def\keyFont{\fontsize{8}{11}\helveticabold }
\def\firstAuthorLast{Chin} 
\def\Authors{Kiyn Chin\,$^{1,2,*}$, Carmel Majidi\,$^{1}$ and Abhinav Gupta$^{2,3}$ }

\def\Address{$^{1}$Soft Machines Lab, Dept. of Mechanical Engineering, Carnegie Mellon University, Pittsburgh, PA, USA \\
$^{2}$Robotics Institute, Carnegie Mellon University, Pittsburgh, PA, USA\\
$^{3}$Meta AI Research, Pittsburgh, PA, USA}

\def\corrAuthor{Kiyn Chin}

\def\corrEmail{kiyn@cmu.edu}

\begin{document}
\onecolumn
\firstpage{1}

\title[Long-Term Soft Robotic Data Collection]{Modular Parallel Manipulator for Long-Term Soft Robotic Data Collection} 

\author[\firstAuthorLast ]{\Authors} 
\address{} 
\correspondance{} 

\extraAuth{}

\maketitle

\begin{abstract}

\section{}
Performing long-term experimentation or large-scale data collection for machine learning in the field of soft robotics is challenging, due to the hardware robustness and experimental flexibility required. In this work, we propose a modular parallel robotic manipulation platform suitable for such large-scale data collection and compatible with various soft-robotic fabrication methods. Considering the computational and theoretical difficulty of replicating the high-fidelity, faster-than-real-time simulations that enable large-scale data collection in rigid robotic systems, a robust soft-robotic hardware platform becomes a high priority development task for the field.

The platform's modules consist of a pair of off-the-shelf electrical motors which actuate a customizable finger consisting of a compliant parallel structure. The parallel mechanism of the finger can be as simple as a single 3D-printed urethane or molded silicone bulk structure, due to the motors being able to fully actuate a passive structure. This design flexibility allows experimentation with soft mechanism varied geometries, bulk properties and surface properties. Additionally, while the parallel mechanism does not require separate electronics or additional parts, these can be included, and it can be constructed using multi-functional soft materials to study compatible soft sensors and actuators in the learning process. In this work, we validate the platform's ability to be used for policy gradient reinforcement learning directly on hardware in a benchmark 2D manipulation task. We additionally demonstrate compatibility with multiple fingers and characterize the design constraints for compatible extensions.

\tiny
 \keyFont{ \section{Keywords:} robotics, soft robotics, soft materials, reinforcement learning}
\end{abstract}

\section{Introduction}

Soft robotic systems are often operated using open-loop control methods that leverage the inherent compliance and conformability of soft materials as they press against a contacting surfaces. Popular examples of this are compliant grippers that can conform to a wide range of objects with limited sensing\cite{king2018}\cite{mannam2021} and soft robots that can passively shape change as they pass through confined spaces \cite{Hawkes2017}.  While material compliance enables intrinsic reactivity  in the form of material deformation upon contact,  there is still  a need for the ability to control the positioning of soft robotic systems in free space, or to specify contact beyond simply modulating force. Computing the deformation of soft materials in isolation can be time intensive. The creation of \textit{A priori} models of soft robotic systems, whose behavior might be dependent on interactions between soft materials with specific, potentially novel compositions or geometries is usually not effective. Therefore, to create models or controllers for soft robotic systems, data-driven methods are quite appealing. Machine learning provides the potential to overcome the challenges of understanding soft material behavior from first principles. However to enable operation of soft material-based robotic systems over longer periods of time, as would be required of a useful tool, the way that soft materials change over time is another reason to move towards the ability to collect larger amounts of hardware data.

\begin{figure}[htbp]
    \centering
    \includegraphics[width=0.7\linewidth]{ 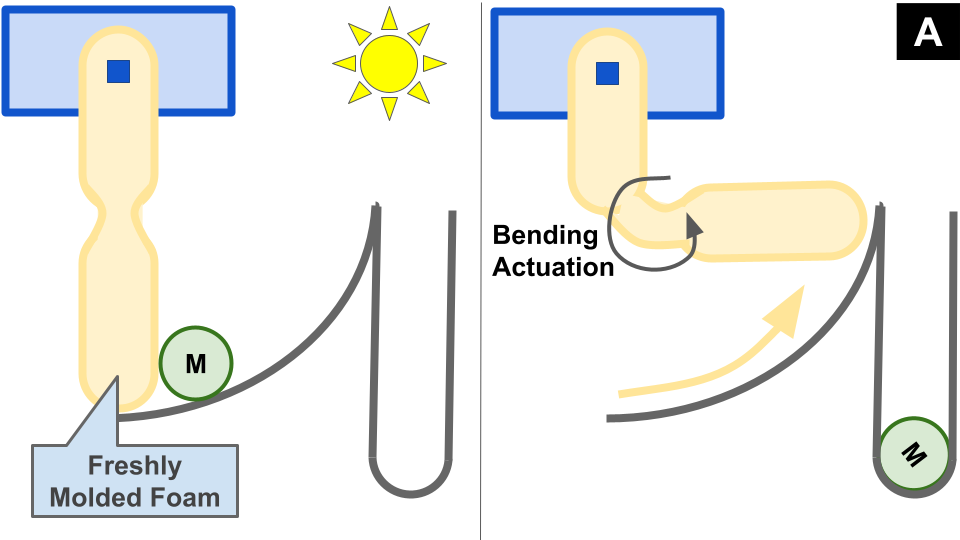}
    \includegraphics[width=0.7\linewidth]{ 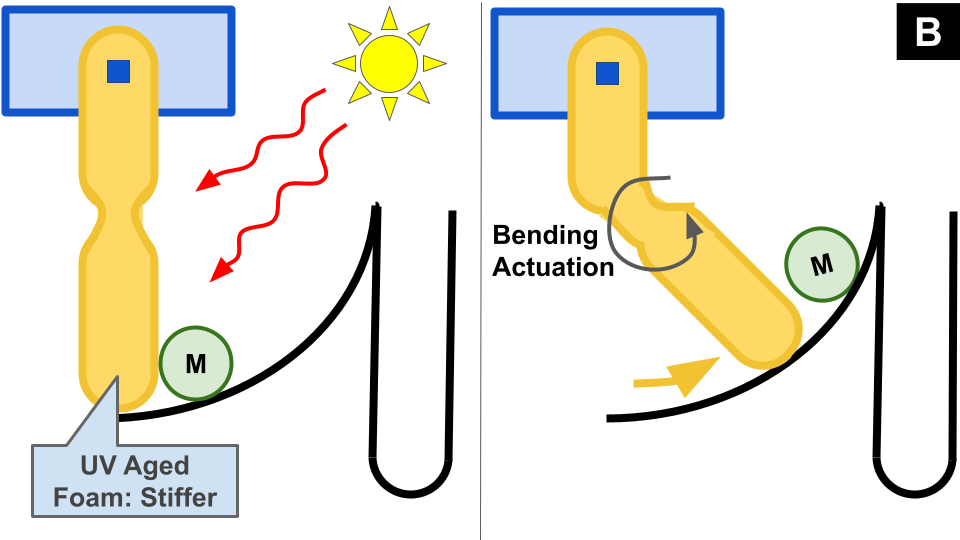}
    \includegraphics[width=0.8\textwidth]{ 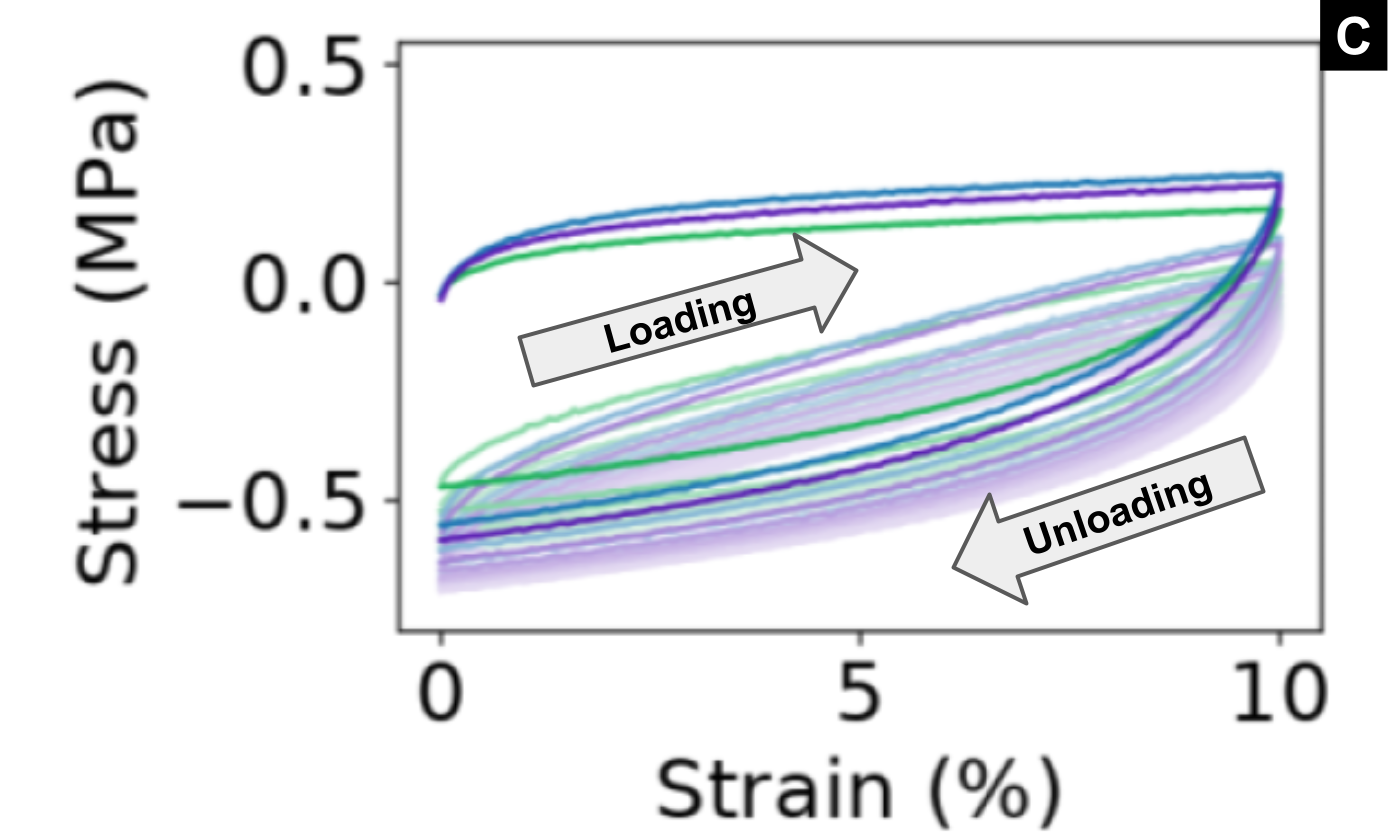}
    \caption[Intrinsic nonstationary behavior of soft materials]{\textbf{A)} Motion of a 1-DOF soft foam manipulator designed to perform a simple open-loop pushing task. \textbf{B)} Eventual task failure due to material aging from sun exposure leading to dynamics drift. \textbf{C)} Intrinsic time-dependent deformation response of soft material to cyclical loading.}  
    \label{fig:TPU}
\end{figure}

While soft robots can be accessible to build, effective control strategies are less clearly available. Soft robot dynamics are difficult to accurately model analytically, due to a multi-physics coupling between shape, forces, physical state (e.g. temperature), and history of motion.  To enable operation of soft robots across longer time scales, it is important to be able to collect hardware data that capture these phenomena, especially those that vary with time. As shown in Fig. \ref{fig:TPU}, the deformation of elastomeric materials has multiple forms of time dependent phenomena. There is hysteresis, which means there is no one-to-one mapping from actuation force to deformation -- instead result of any applied force or deformation depends on the recent history of the system. There is also nonstationarity, which is a distribution shift in the underlying dynamics of a system over time, and can be caused by wear, external temperature fluctuations, internal strain build-up, or many other sources.  Hysteresis and nonstationarity are inherent side-effects of the use of elastomers and other soft materials in the construction of soft robot systems. Popular methods often ignore these effects or treat them as unmodeled noise\cite{Lee2017}\cite{Rolf2015}. Hysteresis can theoretically be addressed by incorporating system state history into the input of models. This explicit time dependence can be encoded in structures like recurrent neural networks\cite{Thuruthel2017} or by simply concatenating multiple time-steps of state data as the input to the model.  Nonstationary behavior necessitates models that can adapt to changing dynamics \cite{Rolf2015}.

These inherent dynamics present an essential problem to the autonomy of soft robotic systems (though deliberate changes to dynamics are an emerging feature of soft robotics \cite{Davis2023}), and may also function as a reasonable proxy for the general problem of imprecise dynamics. Errors in modeling and the drift in models over time exist in rigid-body robotic systems. As described in a report on the problems faced by teams in the DARPA Robotics challenge, on the scale of full humanoid systems, modeling errors due to "wrong kinematic and inertial parameters, cogging and other magnetic effects in electric motors, actuator dynamics and hysteresis, variable viscous or nonlinear dynamic friction depending on the state of hydraulic oil leakage, dust, and dirt, thermal (both external weather and actuator heating) effects, humidity effects, constant delay, variable delay due to computer communications. joint/actuator/transmission stiction and other static friction effects, foot slip-stick on touchdown and during stance, six dimensional bearing play, structural link deformation, and material aging" \cite{atkeson2018} all come into play. More generally, most classes of robotic system can encounter problems with dynamics which do not match the model, either due to mis-specification or nonstationary dynamics.  This is especially true for soft robots and systems composed of mechanically compliant materials.

In traditional control theory, the nonstationarity problem is among those problems addressed by the subfield of adaptive control.  Traditional adaptive control methods often rely on high-fidelity equations of motion with a small number of uncertain parameters\cite{gaudio2019}. There has been work integrating these techniques with neural network models, an approach called concurrent-learning adaptive control (CLAC) \cite{chowdhary2011}. Other approaches have leveraged state-estimation techniques to update analytical parameter sets (Kalman) \cite{kalman2019}.

In the reinforcement learning literature, there have been a few works attempting to solve the problem of learning control under nonstationary dynamics via multiple partial models. These generally train several neural networks for different regions of the drifting dynamics. Multiple Model-based Reinforcement Learning (MMRL) \cite{doya2002} creates a static library of neural models which correspond to a known set of dynamics, and are trained as the system moves through those different modes. Reinforcement Learning with Context Detection (RL-CD)\cite{dasilva2006} trains in a similar way, but rather than a static set of models, there is a context detection module which allows for new models to be created as the system encounters different modes. Hierarchical Reinforcement Learning with Context Detection (HRL-CD)\cite{yucesoy2015} combines the technique of hierarchical reinforcement learning for accelerating learning convergence with the RL-CD framework. All of these techniques make the assumption of deterministic and discrete dynamics.  

There has been work in enabling faster transfer learning for problems in related domains using neural attention and transformer networks \cite{vaswani2017}. Similarly automatic domain randomization (ADR) has been shown to allow for the online formation of reactive controllers which can adapt to a wide range of environmental parameters\cite{Chen_2022randomization, openai2019, Schaff2023}. These methods rely on extremely high-volume data collection for training, requiring high fidelity simulation and compute time on a scale which can be inaccessible to resource-constrained development environments.

Learning for nonstationary dynamics can be seen as an extension of the canonical robot model learning problem. For a system with stationary dynamics, there is the assumption that all experience tuples $(x_t, u_t, x_{t+1})$ are drawn from the same distribution subject to the constraint of the true system dynamics $x_{t+1} = F(x_t,u_t)$, where $x_t$ represents the state of the system at time $t$ and $u_t$ is the control effort at the same time. Since all data provides meaningful, if noisy, information about the true dynamics of the system, all tuples can be used to refine the fidelity of the agents dynamics estimate. In the nonstationary case, this assumption is violated.

For the nonstationarity case, the ground truth system dynamics vary as a function of time. While this variance can be in the form of the dynamics as well as in the values of dynamics parameters, we assume that there exists some parameterization of the variation such that we can write the dynamics as $x_{t+1} = F(x_t,u_t, \psi_t)$, where $\psi_t$ is the parameter vector determining the dynamics at time $t$. 

The behavior of the vector $\psi_t$ is generally unconstrained, and therefore requires domain specific knowledge to create a reasonable set of assumptions. 

Three cases for how the dynamics vector $\psi_t$ might evolve:

\begin{itemize}
    \item Trends in the dynamics,  e.g. material degradation,  polymer creep,  or thermal buildup. These are gradual non-periodic changes.
    \item Random or event-driven changes to dynamics , e.g. mechanism damage, system repair, parts replacement,  or power-cycling. These changes can be unknown or known, and are discrete jumps in the system dynamics.
    \item Cyclical or oscillating dynamics e.g. due to environmental changes corresponding with day-night cycles. These are finite periodic changes
\end{itemize}

\section{Methods}

To better understand the dynamics of soft materials in a long-term experimental environment, we developed a robotic experimental platform which allows automatic performance of a class of simple manipulation tasks using mechanisms made from soft materials. The platform is relatively inexpensive to fabricate and capable of operating for long periods of time. The behavior is  highly coupled to the material properties of the soft materials used in its construction, allowing insights into those properties. We demonstrate the ability to train reinforcement learning policies using only hardware data collected with this system.

\subsection{Motor-driven Soft Parallel Mechanism}
\begin{figure}[htbp]
    \centering
    \begin{tabular}{cc}
       \includegraphics[width=0.4\linewidth]  { 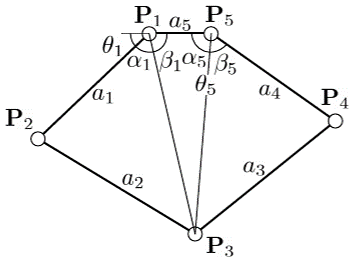} & 
       \includegraphics[width = 0.35\linewidth]{ 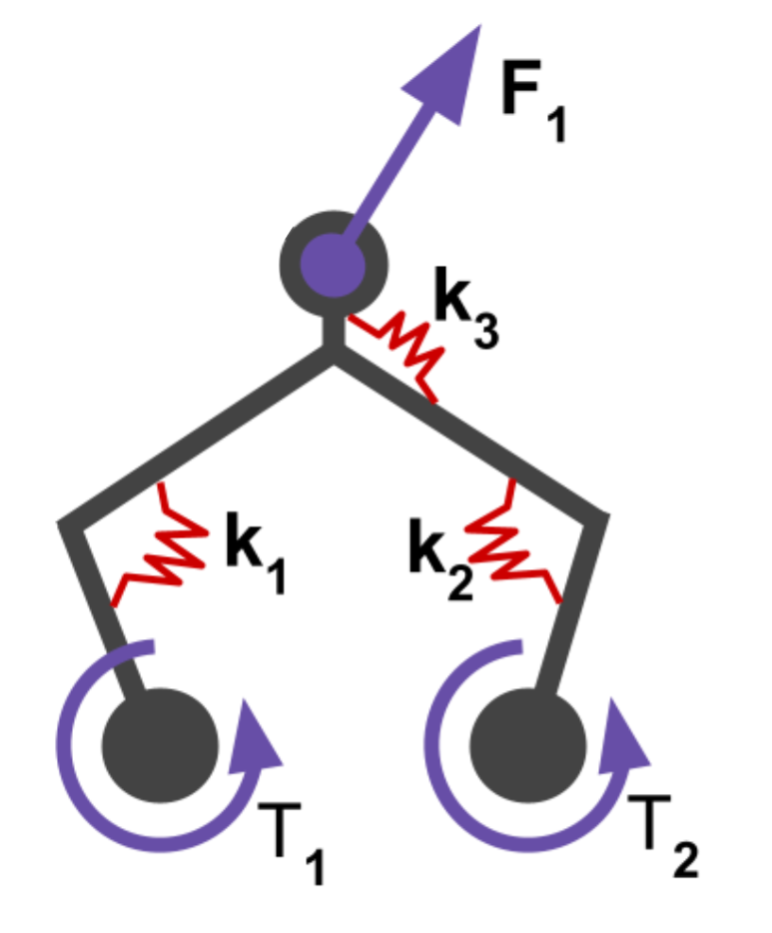}
    \end{tabular}
        
    \caption[Rigid vs. Compliant Five-bar]{Left) Rigid body kinematic approximation of five-bar. Right) Approximate forces of soft material five-bar driven by two servos. T1 and T2 are the actuation space of the system, the torques applied by the servos. Springs $k_1, k_2,$ and $k_3$ are the spring constants of living hinge joints of the five-bar, dependent on hinge material and geometry. The force $F_1$ is the reaction force that occurs when the tip of the five-bar contacts another object. The spring constants of the five-bar links are not shown, but do influence behavior.}
    
    \label{fig:fivebar_kinematics_equilibrium}
\end{figure}

 In order to create a system that can operate for long periods of time during data collection for machine learning or other long-term experiments, we use reliable electric servomotors as the actuators in our design. As many of the moving and interacting parts of the robotic system should be made of soft materials for the properties of those soft materials to be encoded in data of robot motion. Therefore, we employ the electric motors to actuate a soft matter parallel mechanism, a relatively under-explored strategy that has been shown to enable robust, versatile experimental platforms \cite{mannam2021}. The usage of soft materials was focused on the mobile mechanism rather than the whole system because the system components that are not the focus of study should provide as little disturbance to the behavior of the system over time. While soft actuators are diverse, enable many unique properties for robotic systems, and provide a path to fully soft systems, they have relatively complex dynamics. Additionally, the kinds of materials used in their manufacture are not as diverse as for a passive mechanism, and many classes of soft actuator require specialized development and operational environments to maintain \cite{Li2023}. Electric motors are especially robust forms of actuation with minimally impactful actuator dynamics. When these motors are coupled with internal closed loop controllers, this even more the case. We choose to base our designs on planar parallel mechanism, the compliant five-bar.
 
The choice of a five-bar mechanism as the soft structure the robot uses for interaction is helpful partly because of the many ways it can be represented. It is feasible that the geometry of the compliant five-bar could be modeled by discrete elastic rod simulation environment\cite{bergou}. This simulation strategy is one of the fastest found for soft materials, and provides one feature space that could be useful to characterize our system with.  Additionally, the approximation of the compliant five bar via the kinematics of a rigid-body five-bar \ref{fig:fivebar_kinematics_equilibrium}A provides an avenue for the comparison of the effects of the soft materials against a close geometric analogue. The difference in fidelity can be intuitively understood by examining the fact that different thicknesses of joint produce different behavior of a parallel structure, due to changes to spring stiffness (Fig \ref{fig:fivebar_kinematics_equilibrium}), but all joint thicknesses are represented with the same rigid approximation. Parameter estimation of the associated a rigid approximation might provide useful data for adaptation.

\subsection{Compliant Five-bar Modules}

\begin{figure}
    \centering
    \includegraphics[width=0.8\linewidth]{ 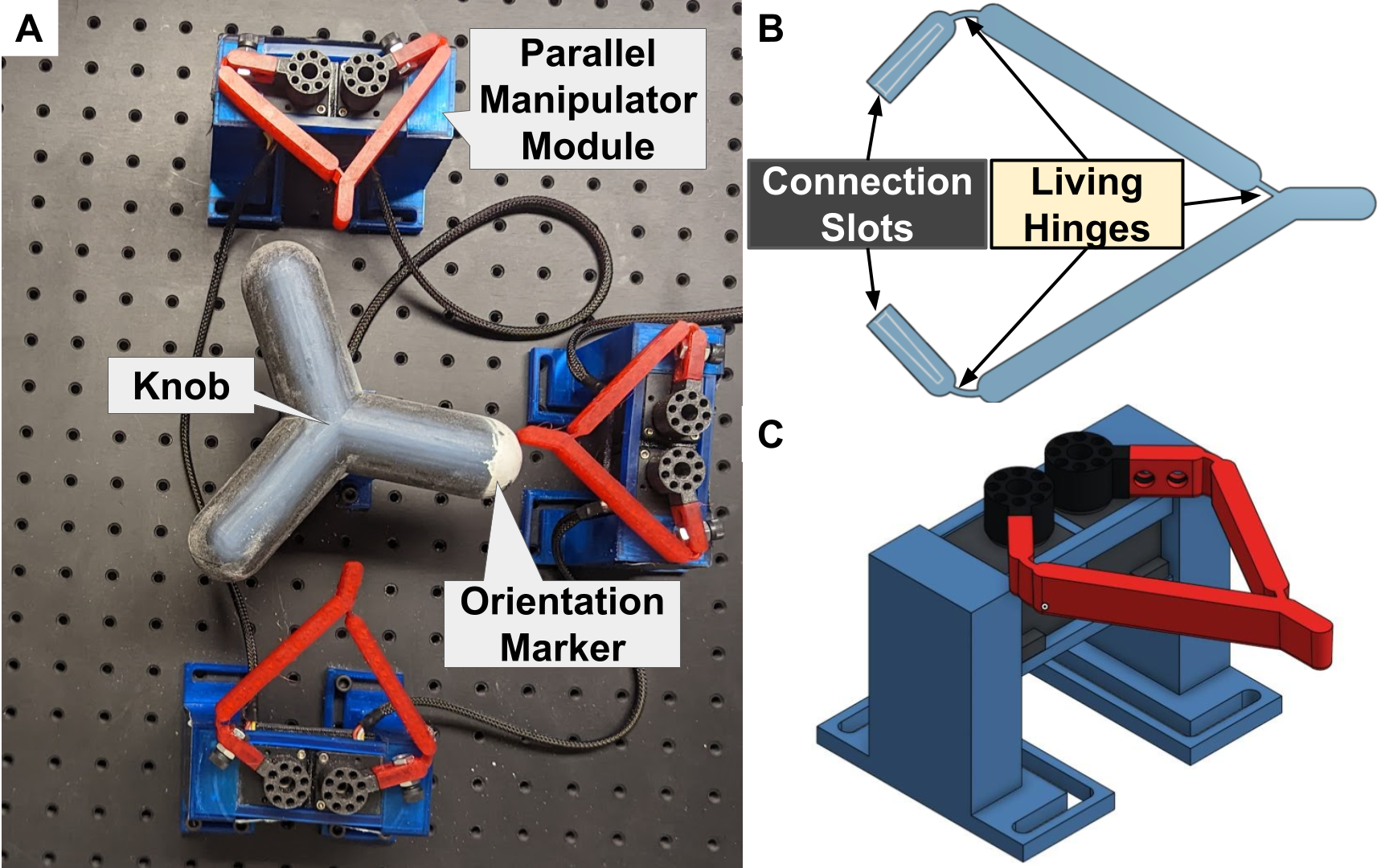}
    \caption[Modular Parallel Manipulator Overview]{A) Full experimental manipulator platform  B) Soft five-bar mechanism design with quick-swap connection slots. C) Individual module design.}
    \label{fig:modular_manipulator_overview}
\end{figure}

\begin{figure}[htbp]
    \centering
    \includegraphics[width=0.8\linewidth]{ 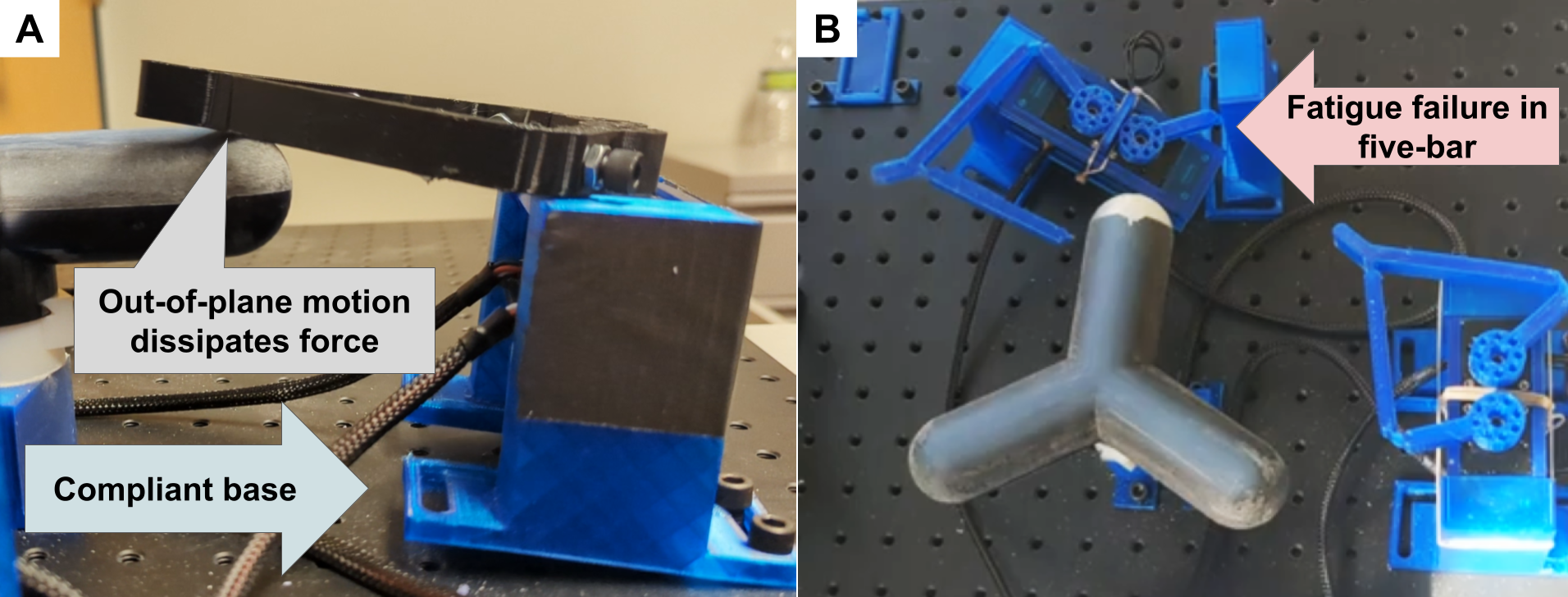}
    \caption{A) By mounting the compliant base of the module to mechanical breadboard using a linear arrangement of bolts, the module is able to rock back, pushing the finger out of plane in response to overly high forces. This reduces the chance of breaking the system during extended unmonitored data collection. B) The most likely failure modes are module disassembly to to friction fit module components and soft five-bar failure, especially at living hinges. Even if both of these occur undetected and the system keeps trying to operate, the compliant materials used and construction lowers the chance of long-term damage to the experimental setup.}
    \label{fig:module_breaking}
\end{figure}

The modules are actuated by two servomotors with aligned axes of rotation mounted next to each other in 3D printed TPU housing, assembled with friction fit \ref{fig:modular_manipulator_overview}. The module design is compatible with both lower cost servomotor (Dynamixel XC430-W150-T, ~\$120), and higher performance servomotor (Dynamixel XM430-W210, ~\$270).The base of the modules are attached to a mechanical breadboard with bolts. These mechanical design choices improve the ability of the system to operate for longer periods of time without requiring repair, as well as simplifying repair \ref{fig:module_breaking}. The primary moving parts of the modules are compliant five-bar mechanisms which mount to the servomotors. The state of an individual module is determined by the angle of the driving servos. There is no autonomously controllable nonstationarity for this system. Instead, the soft components to are the primary source of trends over time in the dynamics. The dynamics can be controlled by swapping out five-bars, which slot onto 3D printed quick-swap servo horns. 

\subsection{Manipulation Task: Knob Turning}

This task is adapted from the ROBEL robotics learning benchmark designed by Ahn et. al \cite{ahn2019robelroboticsbenchmarkslearning}, replacing the rigid-body robot they use with soft five-bar modules. There is a knob-like object in the center of the work space, mounted to a servomotor. The servomotor enables encoder-based state estimation of the knob's pose, and allows the autonomous resetting of the system to a nominal state. This is a critical feature for running machine learning experiments autonomously. This need is also a major motivator of the simple planar geometry chosen for the task. Soft five-bar modules are arrange surrounding the knob, and are tasked with turning the knob to a desired pose. This general task outline can be modified by changing the location of modules, geometry of five-bars, or temporal dynamics of the pose desired, either static goal poses, or pose trajectories. The use of a servo to mount the knob also allows adjustment of the knob's stiffness (as long as the servo used has a torque-control mode).

To maximize the learning speed on hardware, we again leverage dimensionality reduction via action space discretization. There is a relatively natural choice of discrete actions for electric motors in a quasistatic context: increasing or decreasing the angle of the motor by some static amount. These primitives span the space of available states for a motor. This is a superset of the valid states of a module, but by applying geometry-dependent limits to servo position, the full state of each module can be explored by these primitives. The full manipulator array would then have an action space of

\[
 a_t \in \{mag*\Delta \theta (\text{servo}_{i})\}, \quad \text{where} \quad i \in \{0\hdots6\} \quad \text{and} \quad mag \in {\pm \text{angular resolution}}
\]

\begin{figure}[htbp]
    \centering
    \includegraphics[width=\linewidth]{ 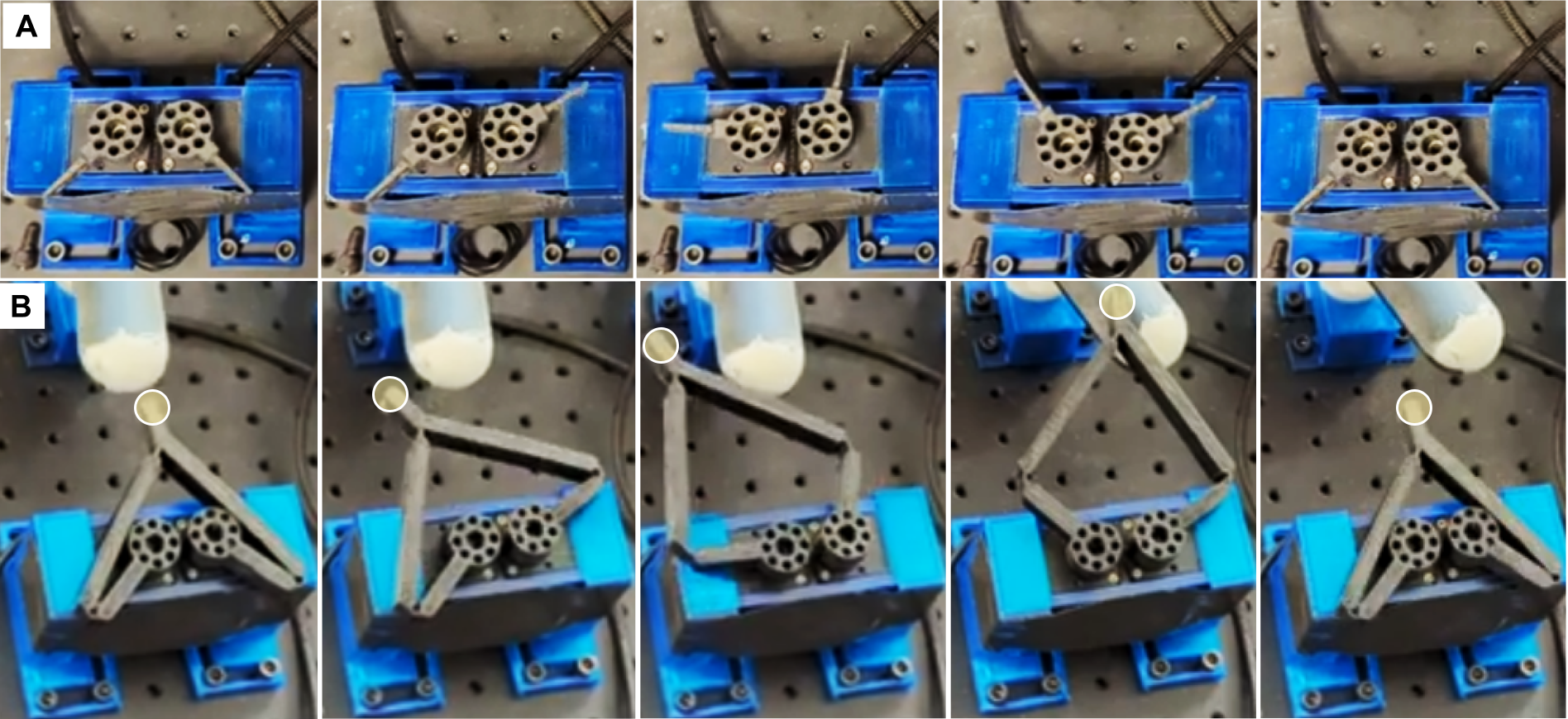}
    \caption{A) Motion primitive execution without attached finger, showing servo horn finger connectors moving. B) Execution of that primitive with a urethane finger mounted to module. }
    \label{fig:manipulator_primitive}
\end{figure}

\noindent However, this low level of abstraction poses some challenges to the speed of training. For studying the ability to bootstrap autonomy, such low level action spaces are often chosen to minimize learned bias and reliance on human design decisions. However, for the goal of making robotic systems development more accessible, there is a different set of considerations. This system is designed to allow more hardware data to be collected than many fully soft robots, yet there is still a time cost of collecting that data. While the easily replaceable modules and resilient mechanical design minimizes damage to the most expensive hardware components, we still aim to minimize the amount of time necessary run hardware to collect data. Therefore, the action space is chosen to maximize the data-efficiency of the learning process, incorporating as much human intuition as possible. First,  an intermediate action space that allows the tip position of each module to be controlled in a "extend", "retract", "left", "right" scheme is developed. Then we build the actual primitives upon this as trajectories of intermediate positions that produce a distinct sweeping motion over several time-steps. Specifically, the motion is a sequence of "move left" $\rightarrow$ "extend" $\rightarrow$ "move right" $\rightarrow$ "retract", shown in Figure \ref{fig:manipulator_primitive}. This sweeping motion and its mirror are able to be performed by each module, resulting in an action space of:

\[
 a_t \in {\text{sweep}(\text{module}_i, \text{dir})}, \quad \text{where} \quad i\in\{0,1,2\}, \quad \text{and} \quad \text{dir} \in \{left,right\}
\]

These primitives are hierarchical, being built first upon the lowest-level motor delta primitives, and intermediate tip-control primitives.  While not explored, this leaves space for learning the dynamics at a lower level more optimal or less biased behavior is desired later on. These sweeping primitives are very likely to move the knob if the moving module is aligned with a lobe of the knob geometry. This means that the learning problem is figuring out how to sequence the primitives to complete the task, requiring the learning of some latent representation of the relative geometries of the modules and the knob, and the way the modules interact when making contact with the knob, but not needing to learn coherent motion. 

For the following results, the task is learning to turn the knob to a defined goal position from a defined start position. 

\section{Results}
\subsection{Reinforcement Learning on Hardware}

Reinforcement learning was used to learn this task from empirical hardware data. We used policy gradient methods from the Stable Baselines3 repository of standard reinforcement learning algorithms \cite{stable_baselines3}, specifically Proximal Policy Optimization (PPO) with a batch size of 64 and a learning rate of $1\text{e}^{-3}$. We wrapped the hardware controller for the modules in a Gym environment.

The action space of the reinforcement learning problem was the sweeping primitives derived in the previous section. The state space is the servo angles of each module, normalized between -1 and 1.  Appended to this is the knob pose, represented as a two-element vector of the cosine of knob angle and the sine of the knob angle, for better numerical performance in the policy network. The reward used for reinforcement learning was

\[
\text{Reward} = -5 \times \left| \Delta\theta_{\text{knob}} \right| 
+ 10 \times \mathbb{1}\left\{ \left| \Delta\theta_{\text{knob}} \right| \geq \frac{0.25}{\pi} \right\} 
+ 50 \times \mathbb{1}\left\{ \left| \Delta\theta_{\text{knob}} \right| \geq \frac{0.10}{\pi} \right\}, 
 \]

 \noindent corresponding to a smoothly increasing cost of 5 times the angle of the knob in radians, penalizing being farther away from the goal position of 0 radians. There is a big reward jump of 10 if within $\frac{0.25}{\pi}$ radians, and an even bigger reward of 50 if within $\frac{0.10}{\pi}$ radians of the goal position.

The reinforcement learning experiments were run in increasing lengths of time, following a powers of two progression. The first experiment was $2^{14} =  16, 284 $ time-steps and progressed up to $2^{16}=65,536$ time steps. The next step up, $2^{17} = 131,072$ time-steps, was also run, but it was partway into this that the system finally experienced a hardware failure in the form of one soft finger failing at a hinge and pushing itself out of its housing, like the failure mode shown in Figure \ref{fig:module_breaking}B. The repair process for such a failure was rather quick, requiring only the attachment of a new finger with bolts to the quick-swap servo horn, and the fastener-free reassembly of the module. 

The length of a time step is the amount of time it takes to execute a primitive, which is about 5 seconds. This means that the longest completed continuous batch of data collection, the ~65k batch, ran for 3.76 days straight. In all, counting from the time the system was activated, all the completed batches as well as the 10,714 time-steps completed in the 131k batch before hardware failure, the system was active for $2^{14}+2^{15}+2^{16}+10,714 = 125,402$ time-steps, or 7.25 days. There were automated cooldown periods between batches, totalling to 2 hours of downtime, meaning a total 7.34 days from the beginning of the data collection.  This means the system operated for over a week at 98.7\% uptime before requiring repair. 

\begin{figure} [htbp]
    \centering    \includegraphics[width=\linewidth]{ 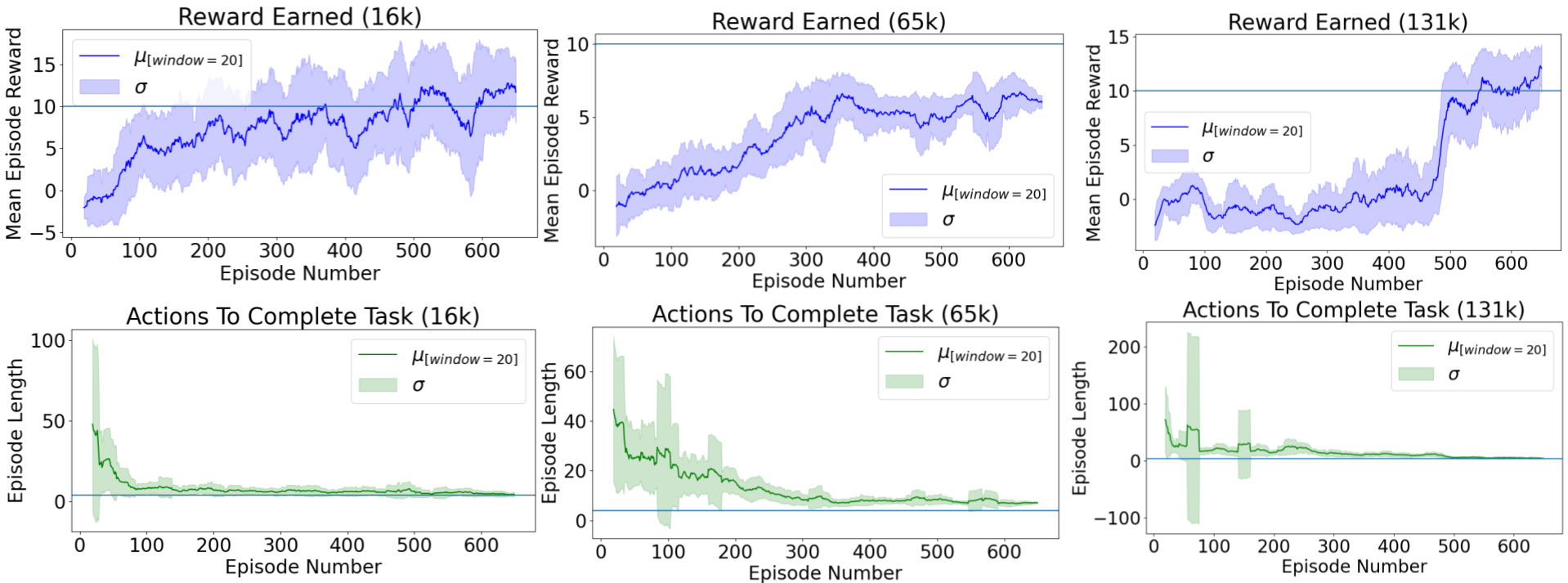}
    \caption{Policy improvement for different training sessions. These traces are extracted from the first 650 episodes of each training session. The reward value of 10 and episode length of 4 are marked off as they roughly correspond to the best performance found throughout experimentation.}
\label{fig:training_trajectories}
\end{figure}

\begin{figure}[htbp]
    \centering
    \includegraphics[width=\linewidth]{ 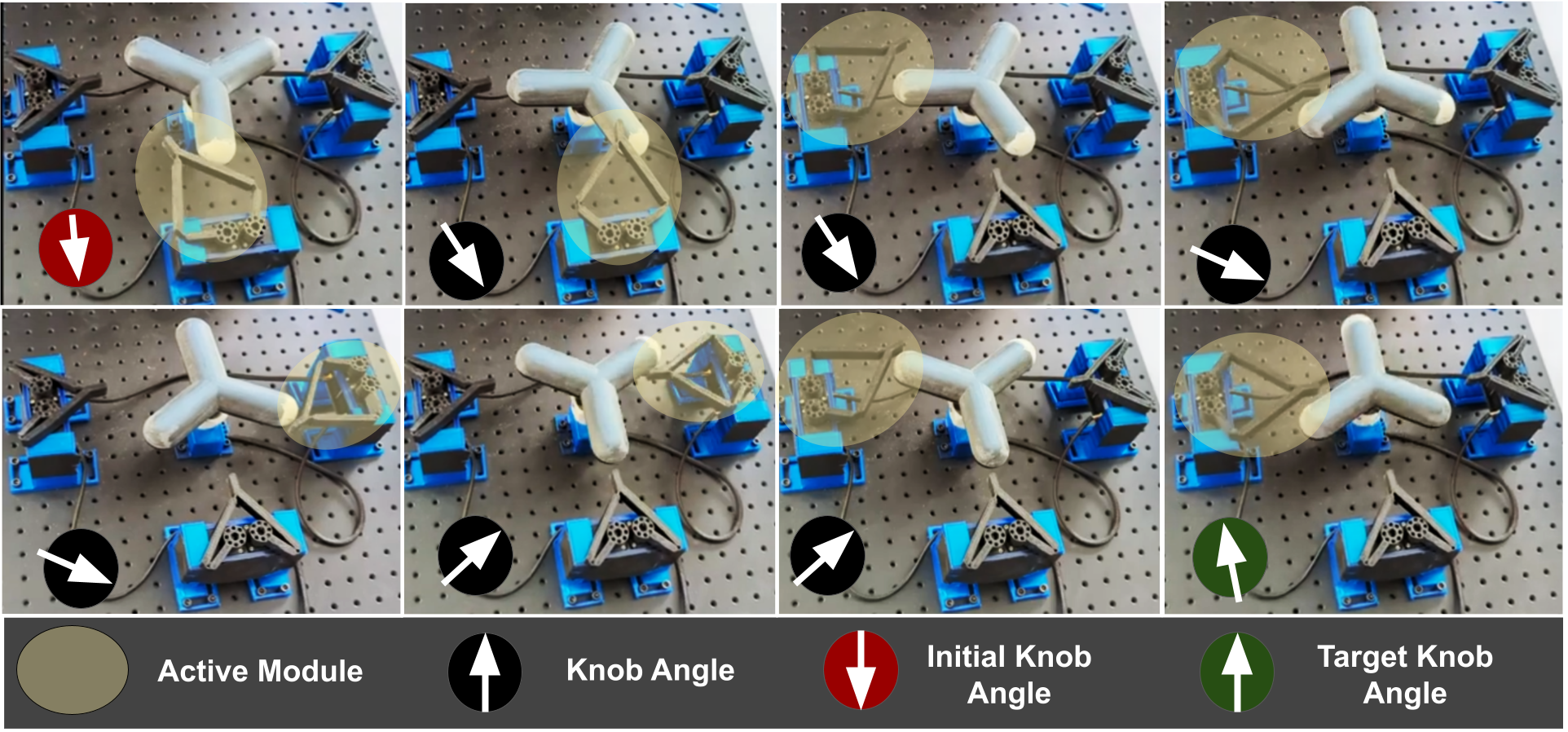}
    \caption{Successful policy execution leading to completion knob turning task using urethane five-bars. Policy learned via policy gradient reinforcement learning over action primitives. }
    \label{fig:manipulator_task_completion}
\end{figure}

The system was able to improve in task performance over time, as shown in Figure \ref{fig:training_trajectories}. Based on the reward function, achieving a mean reward of 10 was significant as it requires quickly moving the system close to the goal without spending many time steps. The episode length of 4 is significant as that was almost the fastest possible execution (shown in Figure \ref{fig:manipulator_task_completion}. Episode lengths of 3 were observed, but they involved hard-to-replicate interactions, where due to the angle of contact, elastic energy was stored in the five-bar before being released and pushing the knob further than the quasistatic sweeping motion.

\begin{figure} [htbp]
    \centering
    \includegraphics[width=0.8\linewidth]{ 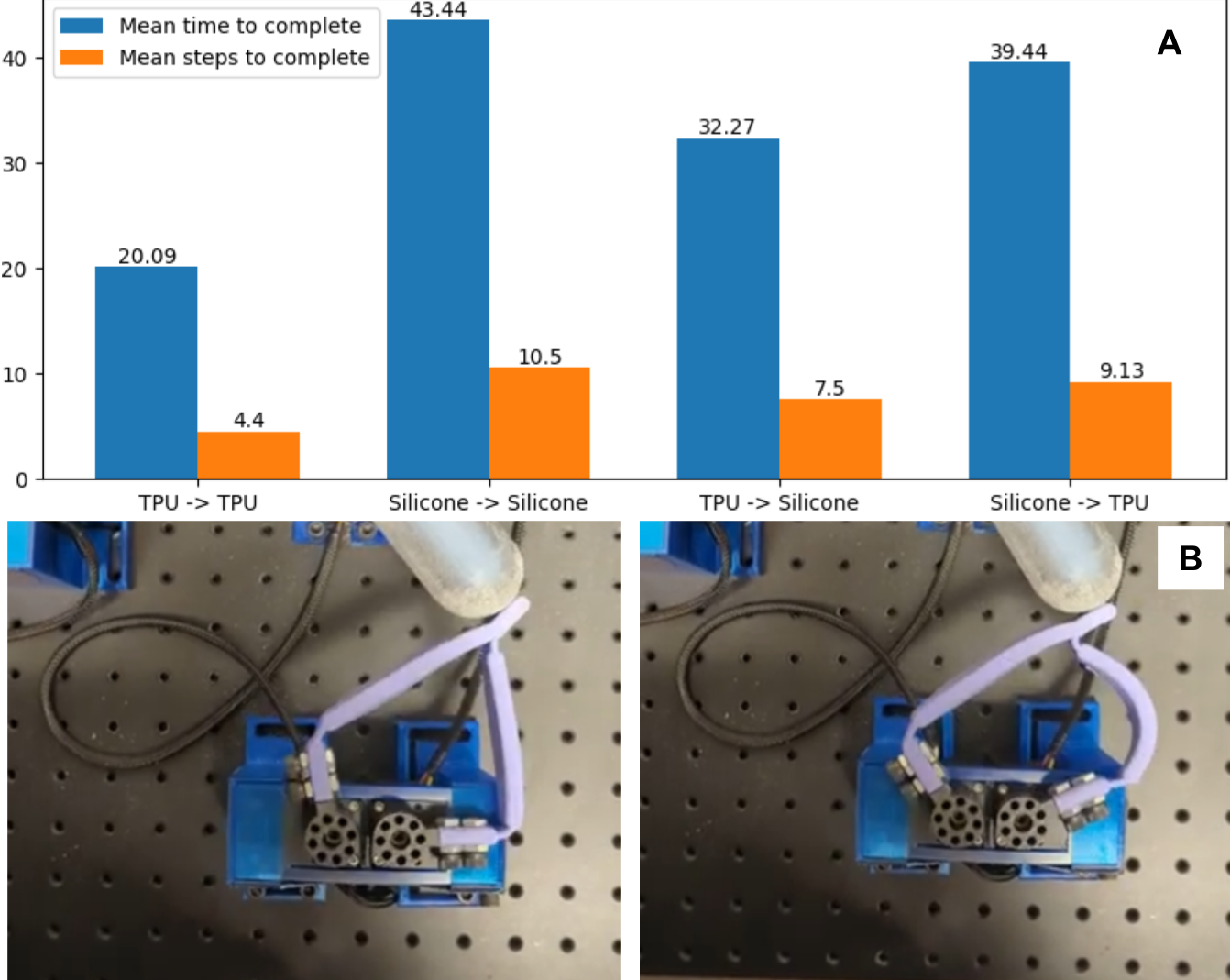}
    \caption{A) Performance comparison of transferred policies, trained on different soft materials. Each case tested 10 times. B) Significant bending of links in softer silicone indicates greater deviation from nominal kinematics.}
    \label{fig:crosstesting_data}
\end{figure}

Once the ability to collect enough hardware data to perform reinforcement learning was demonstrated, we evaluated the platform's ability to provide insight into the effects of different soft materials on the dynamics. 

\subsection{Policy Transfer Between Soft Materials}
\begin{figure}[htbp]
    \centering
    \includegraphics[width=0.8\linewidth]{ 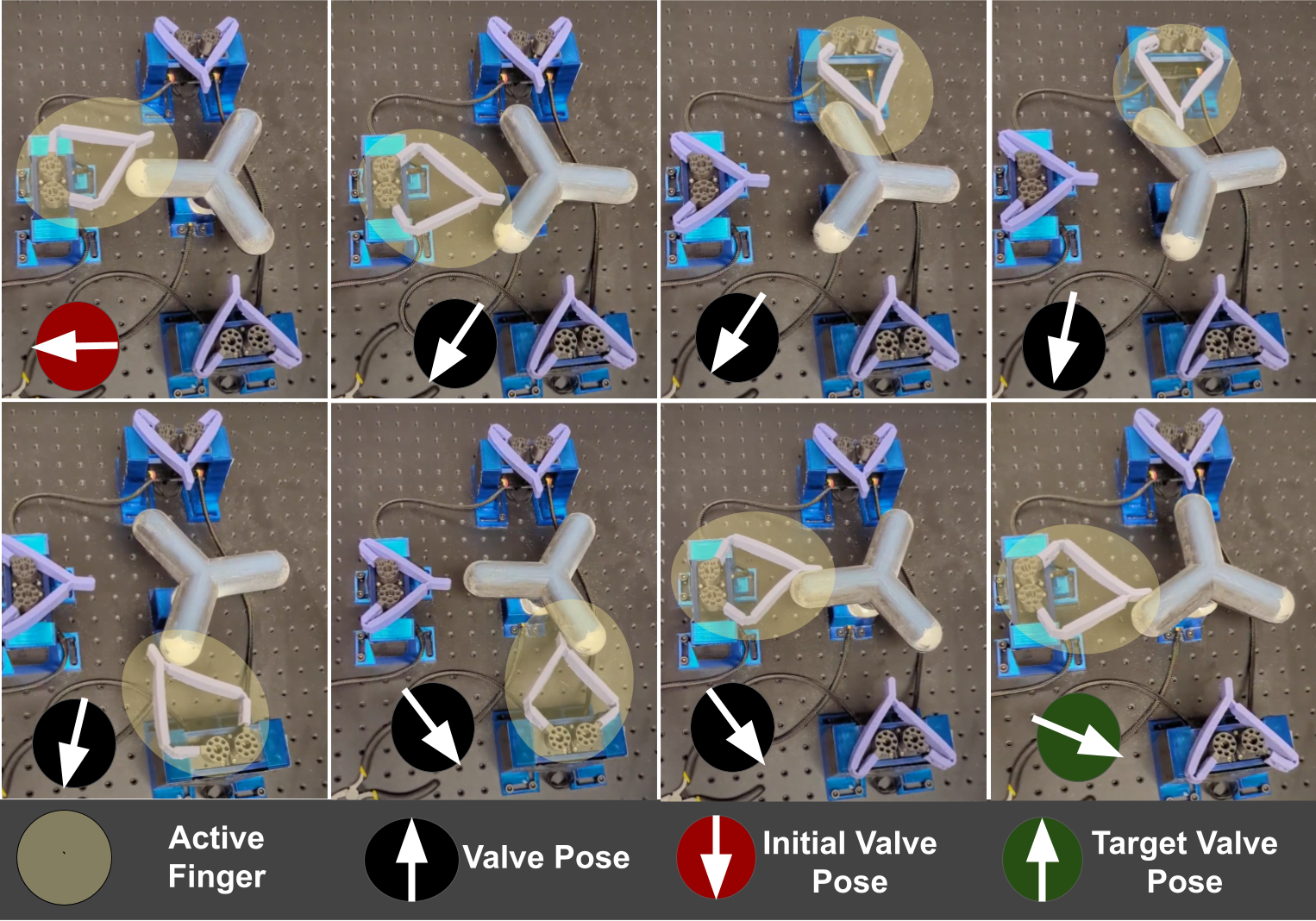}
    \caption{Successful policy transfer of knob turning task from urethane five-bars to silicone five-bars.}
    \label{fig:silicone_task_completion}
\end{figure}

Previous results were obtained using 3D printed thermoplastic urethane five-bars. We created new compliant five-bars from injection-molded silicone (Ecoflex 00-30, Smooth-On), a much softer material. These were manufactured with a desktop injection molding setup, powered by compressed air. We trained the system to perform the knob turning task using the reinforcement learning system outlined above. In Figure \ref{fig:silicone_task_completion}, we can see the execution of this policy on hardware.

We then characterized the performance of the policy learned on the silicone five-bars when executed on silicone-five bars as well as on TPU five-bars. We also characterized the policy learned on the TPU five-bars to both systems. The results of these experiments are shown in Figure \ref{fig:crosstesting_data}A.

The best performing case was TPU-trained policy executed on TPU. It might be expected that the dominant factor in transfer performance is similarity between system dynamics during training and system dynamics during execution. However, we see that the next best performing case is a TPU-trained policy executed on silicone. This suggests that the material properties of TPU are more efficient for learning. This makes sense as the softness of silicone means that deformation of the five-bar links outside of the living hinges are more significant, as shown in Figure \ref{fig:crosstesting_data}B. This provides more possible physical configurations for a given motor position. This effectively increases the noisiness of state transitions, meaning learning is a longer process. Additionally, the worst performing case is a silicone-trained policy on a silicone five-bar. The fact that a policy trained on silicone performs better on TPU indicates that the stiffness of TPU is a boon for executing actions reliably.

The ability to extract insights about the suitability of different materials for this task form simple experiments like this supports the utility of this kind of research platform for studying the effect of soft material dynamics on robotic system operation. 

\section{Acknowledgements}
Thank you to Tess Hellebrekers for fabrication of knob hardware, to Richard Desatnik for fabrication of silicone five-bars. Thank you to Vikash Kumar, for firmware optimization for the Dynamixel servos and introduction to ROBEL task.

\section*{Conflict of Interest Statement}

The authors declare that the research was conducted in the absence of any commercial or financial relationships that could be construed as a potential conflict of interest.

\section*{Data Availability Statement}
The code for this study can be found in the following \href{Github repository}{https://github.com/kiynchin/module-nonstationarity}.

\bibliographystyle{unsrtnat} 
\bibliography{Ref}

\begin{thebibliography}{22}
\providecommand{\natexlab}[1]{#1}
\providecommand{\url}[1]{\texttt{#1}}
\expandafter\ifx\csname urlstyle\endcsname\relax
  \providecommand{\doi}[1]{doi: #1}\else
  \providecommand{\doi}{doi: \begingroup \urlstyle{rm}\Url}\fi

\bibitem[King et~al.(2019)King, Bauer, Schlagenhauf, Chang, Moro, Pollard, and
  Coros]{king2018}
Jonathan~P. King, Dominik Bauer, Cornelia Schlagenhauf, Kai~Hung Chang, Daniele
  Moro, Nancy Pollard, and Stelian Coros.
\newblock Design. fabrication, and evaluation of tendon-driven multi-fingered
  foam hands.
\newblock \emph{IEEE-RAS International Conference on Humanoid Robots},
  2018-November:\penalty0 540--545, 1 2019.
\newblock \doi{10.1109/HUMANOIDS.2018.8624997}.

\bibitem[Mannam et~al.(2021)Mannam, Rudich, Zhang, Veloso, Kroemer, and
  Temel]{mannam2021}
Pragna Mannam, Avi Rudich, Kevin Zhang, Manuela Veloso, Oliver Kroemer, and
  F~Zeynep Temel.
\newblock A low-cost compliant gripper using cooperative mini-delta robots for
  dexterous manipulation.
\newblock \emph{Robotics Science and Systems}, 2021.
\newblock URL \url{https://sites.google.com/view/mini-delta-robots}.

\bibitem[Hawkes et~al.(2017)Hawkes, Blumenschein, Greer, and
  Okamura]{Hawkes2017}
Elliot~W. Hawkes, Laura~H. Blumenschein, Joseph~D. Greer, and Allison~M.
  Okamura.
\newblock A soft robot that navigates its environment through growth.
\newblock \emph{Science Robotics}, 2\penalty0 (8):\penalty0 eaan3028, 2017.
\newblock \doi{10.1126/scirobotics.aan3028}.
\newblock URL
  \url{https://www.science.org/doi/abs/10.1126/scirobotics.aan3028}.

\bibitem[Lee et~al.(2017)Lee, Kim, Kim, Hong, Ryu, Kim, and Kim]{Lee2017}
Chiwon Lee, Myungjoon Kim, Yoon~Jae Kim, Nhayoung Hong, Seungwan Ryu, H.~Jin
  Kim, and Sungwan Kim.
\newblock {Soft robot review}.
\newblock \emph{International Journal of Control, Automation and Systems},
  15\penalty0 (1):\penalty0 3--15, 2017.
\newblock ISSN 20054092.
\newblock \doi{10.1007/s12555-016-0462-3}.

\bibitem[Rolf et~al.(2015)Rolf, Neumann, Quei{\ss}er, Reinhart, Nordmann, and
  Steil]{Rolf2015}
M.~Rolf, K.~Neumann, J.~F. Quei{\ss}er, R.~F. Reinhart, A.~Nordmann, and J.~J.
  Steil.
\newblock {A multi-level control architecture for the bionic handling
  assistant}.
\newblock \emph{Advanced Robotics}, 29\penalty0 (13):\penalty0 847--859, 7
  2015.
\newblock ISSN 15685535.
\newblock \doi{10.1080/01691864.2015.1037793}.

\bibitem[Thuruthel et~al.(2017)Thuruthel, Falotico, Renda, and
  Laschi]{Thuruthel2017}
Thomas~George Thuruthel, Egidio Falotico, Federico Renda, and Cecilia Laschi.
\newblock {Learning dynamic models for open loop predictive control of soft
  robotic manipulators}.
\newblock \emph{Bioinspiration and Biomimetics}, 12\penalty0 (6), 2017.
\newblock ISSN 17483190.
\newblock \doi{10.1088/1748-3190/aa839f}.

\bibitem[Davis et~al.(2023)Davis, Woodman, Landesberg, Kramer-Bottiglio, and
  Bongard]{Davis2023}
Q.~Tyrell Davis, Stephanie Woodman, Melanie Landesberg, Rebecca
  Kramer-Bottiglio, and Josh Bongard.
\newblock Subtract to adapt: Autotomic robots.
\newblock In \emph{2023 IEEE International Conference on Soft Robotics
  (RoboSoft)}, pages 1--6, 2023.
\newblock \doi{10.1109/RoboSoft55895.2023.10122102}.

\bibitem[Atkeson et~al.(2018)Atkeson, Wisely~Babu, Banerjee, Berenson, Bove,
  Cui, Dedonato, Du, Feng, Franklin, Gennert, Graff, He, Jaeger, Kim, Knoedler,
  Li, Liu, Long, and Xinjilefu]{atkeson2018}
Christopher Atkeson, Benzun~Pious Wisely~Babu, Nandan Banerjee, Dmitry
  Berenson, Christoper Bove, Xiongyi Cui, Mathew Dedonato, Ruixiang Du, Siyuan
  Feng, Perry Franklin, M.~Gennert, Joshua Graff, Peng He, Aaron Jaeger,
  Joohyung Kim, Kevin Knoedler, Lening Li, Chenggang Liu, Xianchao Long, and
  X.~Xinjilefu.
\newblock \emph{What Happened at the DARPA Robotics Challenge Finals}, pages
  667--684.
\newblock Springer, 04 2018.
\newblock ISBN 978-3-319-74665-4.
\newblock \doi{10.1007/978-3-319-74666-1_17}.

\bibitem[Gaudio et~al.(2019)Gaudio, Gibson, Annaswamy, Bolender, and
  Lavretsky]{gaudio2019}
Joseph~E Gaudio, Travis~E Gibson, Anuradha~M Annaswamy, Michael~A Bolender, and
  Eugene Lavretsky.
\newblock Connections between adaptive control and optimization in machine
  learning.
\newblock In \emph{2019 IEEE 58th Conference on Decision and Control (CDC)},
  2019.

\bibitem[Chowdhary and Johnson(2011)]{chowdhary2011}
Girish~V. Chowdhary and Eric~N. Johnson.
\newblock Theory and flight-test validation of a concurrent-learning adaptive
  controller.
\newblock \emph{Journal of Guidance, Control, and Dynamics}, 34:\penalty0
  592--607, 5 2011.
\newblock ISSN 15333884.
\newblock \doi{10.2514/1.46866}.
\newblock URL \url{https://arc.aiaa.org/doi/abs/10.2514/1.46866}.

\bibitem[Abuduweili and Liu(2019)]{kalman2019}
Abulikemu Abuduweili and Changliu Liu.
\newblock Robust online model adaptation by extended kalman filter with
  exponential moving average and dynamic multi-epoch strategy.
\newblock \emph{arXiv}, 120:\penalty0 1--14, 12 2019.
\newblock URL \url{http://arxiv.org/abs/1912.01790}.

\bibitem[K et~al.(2002)K, K, K, and M]{doya2002}
Doya K, Samejima K, Katagiri K, and Kawato M.
\newblock Multiple model-based reinforcement learning.
\newblock \emph{Neural computation}, 14:\penalty0 1347--1369, 6 2002.
\newblock ISSN 0899-7667.
\newblock \doi{10.1162/089976602753712972}.
\newblock URL \url{https://pubmed.ncbi.nlm.nih.gov/12020450/}.

\bibitem[Silva et~al.(2006)Silva, Basso, Bazzan, and Engel]{dasilva2006}
Bruno C~Da Silva, Eduardo~W Basso, Ana L~C Bazzan, and Paulo~M Engel.
\newblock Dealing with non-stationary environments using context detection.
\newblock In \emph{ICML '06: Proceedings of the 23rd international conference
  on Machine learning}, 2006.

\bibitem[Yücesoy and Tümer(2015)]{yucesoy2015}
Yiğit E.~Yücesoy Yücesoy and M.~Borahan Tümer.
\newblock Hierarchical reinforcement learning with context detection (hrl-cd).
\newblock \emph{International Journal of Machine Learning and Computing},
  5:\penalty0 353--358, 10 2015.
\newblock ISSN 20103700.
\newblock \doi{10.7763/ijmlc.2015.v5.533}.

\bibitem[Vaswani et~al.(2017)Vaswani, Shazeer, Parmar, Uszkoreit, Jones, Gomez,
  Kaiser, and Polosukhin]{vaswani2017}
Ashish Vaswani, Noam Shazeer, Niki Parmar, Jakob Uszkoreit, Llion Jones,
  Aidan~N. Gomez, \L{}ukasz Kaiser, and Illia Polosukhin.
\newblock Attention is all you need.
\newblock In \emph{Proceedings of the 31st International Conference on Neural
  Information Processing Systems}, NIPS'17, page 6000–6010, Red Hook, NY,
  USA, 2017. Curran Associates Inc.
\newblock ISBN 9781510860964.

\bibitem[Chen et~al.(2022)Chen, Hu, Jin, Li, and Wang]{Chen_2022randomization}
Xiaoyu Chen, Jiachen Hu, Chi Jin, Lihong Li, and Liwei Wang.
\newblock Understanding domain randomization for sim-to-real transfer.
\newblock In \emph{10th International Conference on Learning Representations,
  ICLR 2022}, 2022.
\newblock Publisher Copyright: {\textcopyright} 2022 ICLR 2022 - 10th
  International Conference on Learning Representationss. All rights reserved.;
  10th International Conference on Learning Representations, ICLR 2022 ;
  Conference date: 25-04-2022 Through 29-04-2022.

\bibitem[OpenAI et~al.(2019)OpenAI, Akkaya, Andrychowicz, Chociej, Litwin,
  McGrew, Petron, Paino, Plappert, Powell, Ribas, Schneider, Tezak, Tworek,
  Welinder, Weng, Yuan, Zaremba, and Zhang]{openai2019}
OpenAI, Ilge Akkaya, Marcin Andrychowicz, Maciek Chociej, Mateusz Litwin, Bob
  McGrew, Arthur Petron, Alex Paino, Matthias Plappert, Glenn Powell, Raphael
  Ribas, Jonas Schneider, Nikolas Tezak, Jerry Tworek, Peter Welinder, Lilian
  Weng, Qiming Yuan, Wojciech Zaremba, and Lei Zhang.
\newblock Solving rubik's cube with a robot hand.
\newblock \emph{arXiv}, 10 2019.
\newblock URL \url{https://arxiv.org/abs/1910.07113v1}.

\bibitem[Schaff et~al.(2023)Schaff, Sedal, Ni, and Walter]{Schaff2023}
Charles Schaff, Audrey Sedal, Shiyao Ni, and Matthew~R. Walter.
\newblock Sim-to-real transfer of co-optimized soft robot crawlers.
\newblock \emph{Autonomous Robots}, 47\penalty0 (8):\penalty0 1195--1211, 2023.
\newblock \doi{10.1007/s10514-023-10130-8}.
\newblock URL \url{https://doi.org/10.1007/s10514-023-10130-8}.

\bibitem[Li et~al.(2023)Li, Hu, and Yang]{Li2023}
Weidong Li, Diangang Hu, and Lei Yang.
\newblock Actuation mechanisms and applications for soft robots: A
  comprehensive review.
\newblock \emph{Applied Sciences}, 13\penalty0 (16), 2023.
\newblock ISSN 2076-3417.
\newblock \doi{10.3390/app13169255}.
\newblock URL \url{https://www.mdpi.com/2076-3417/13/16/9255}.

\bibitem[Bergou et~al.(2008)Bergou, Wardetzky, Robinson, and Audoly]{bergou}
Miklós Bergou, Max Wardetzky, Stephen Robinson, and Basile Audoly.
\newblock Discrete elastic rods.
\newblock \emph{SIGGRAPH}, 2008.

\bibitem[Ahn et~al.(2019)Ahn, Zhu, Hartikainen, Ponte, Gupta, Levine, and
  Kumar]{ahn2019robelroboticsbenchmarkslearning}
Michael Ahn, Henry Zhu, Kristian Hartikainen, Hugo Ponte, Abhishek Gupta,
  Sergey Levine, and Vikash Kumar.
\newblock Robel: Robotics benchmarks for learning with low-cost robots, 2019.
\newblock URL \url{https://arxiv.org/abs/1909.11639}.

\bibitem[sta(2024)]{stable_baselines3}
2024.
\newblock URL \url{https://stable-baselines3.readthedocs.io/en/master/}.

\end{thebibliography}

\end{document}